\documentclass{article}
\usepackage{acra}
\usepackage{graphicx}
\usepackage{url}

\usepackage{enumitem}
\usepackage[acronyms, shortcuts]{glossaries}
\usepackage{xcolor}
\usepackage{subcaption}
\usepackage{adjustbox}
\usepackage{comment}
\usepackage{balance}

\newacronym[longplural={Bird's Eye View}]{bev}{BEV}{Bird's Eye View}
\newacronym[longplural={convolutional neural networks}]{cnn}{CNN}{convolutional neural network}
\newacronym[longplural={Vision Transformers}]{vit}{ViT}{Vision Transformer}
\newacronym[longplural={Graphics Processing Units}]{gpu}{GPU}{Graphics Processing Unit}
\newacronym[longplural={artificial intelligence}]{ai}{AI}{artificial intelligence}
\newacronym[longplural={intersection-over-unions}]{iou}{IoU}{intersection-over-union}
\newacronym[longplural={mean intersection-over-union}]{miou}{mIoU}{mean intersection-over-union}   

\title{Few-shot Semantic Learning for Robust Multi-Biome 3D Semantic Mapping in Off-Road Environments}

\author{Deegan Atha\textsuperscript{1}\thanks{This work was carried out at the Jet Propulsion Laboratory, California Institute of Technology, under a contract with the National Aeronautics and Space Administration (80NM0018D0004). This work was partially supported by Defense Advanced Research Projects Agency (DARPA). Approved for Public Release, Distribution Unlimited.}\thanks{\copyright 2024. California Institute of Technology. Government sponsorship acknowledged. All rights reserved.}, Xianmei Lei\textsuperscript{1},  Shehryar Khattak\textsuperscript{1}, Anna Sabel\textsuperscript{1}, Elle Miller\textsuperscript{1,2}, \\
\textbf{Aurelio Noca\textsuperscript{1,3}, Grace Lim\textsuperscript{1}, Jeffrey Edlund\textsuperscript{1}, Curtis Padgett\textsuperscript{1}, Patrick Spieler\textsuperscript{1}}\\
\textsuperscript{1} Jet Propulsion Laboratory, California Institute of Technology, United States \\
\textsuperscript{2} The University of Sydney, Australia \\
\textsuperscript{3} Swiss Federal Institute of Technology Lausanne, Switzerland \\
\{firstname.lastname\}@jpl.nasa.gov}

\makeatletter
\def\thanks#1{\protected@xdef\@thanks{\@thanks
        \protect\footnotetext{#1}}}
\makeatother

\begin{document}

\maketitle              

\begin{abstract}
Off-road environments pose significant perception challenges for high-speed autonomous navigation due to unstructured terrain, degraded sensing conditions, and domain-shifts among biomes. 
Learning semantic information across these conditions and biomes can be challenging when a large amount of ground truth data is required.
In this work, we propose an approach that leverages a pre-trained \ac{vit} with fine-tuning on a small ($<$500 images), sparse and coarsely labeled ($<$30\% pixels) multi-biome dataset to predict 2D semantic segmentation classes. These classes are fused over time via a novel range-based metric and aggregated into a 3D semantic voxel map.
We demonstrate zero-shot out-of-biome 2D semantic segmentation on the Yamaha (52.9 mIoU) and Rellis (55.5 mIoU) datasets along with few-shot coarse sparse labeling with existing data for improved segmentation performance on Yamaha (66.6 mIoU) and Rellis (67.2 mIoU).
We further illustrate the feasibility of using a voxel map with a range-based semantic fusion approach to handle common off-road hazards like pop-up hazards, overhangs, and water features.
\end{abstract}
\section{Introduction}

\begin{figure*}[t]
    \centering
    \includegraphics[width=\textwidth]{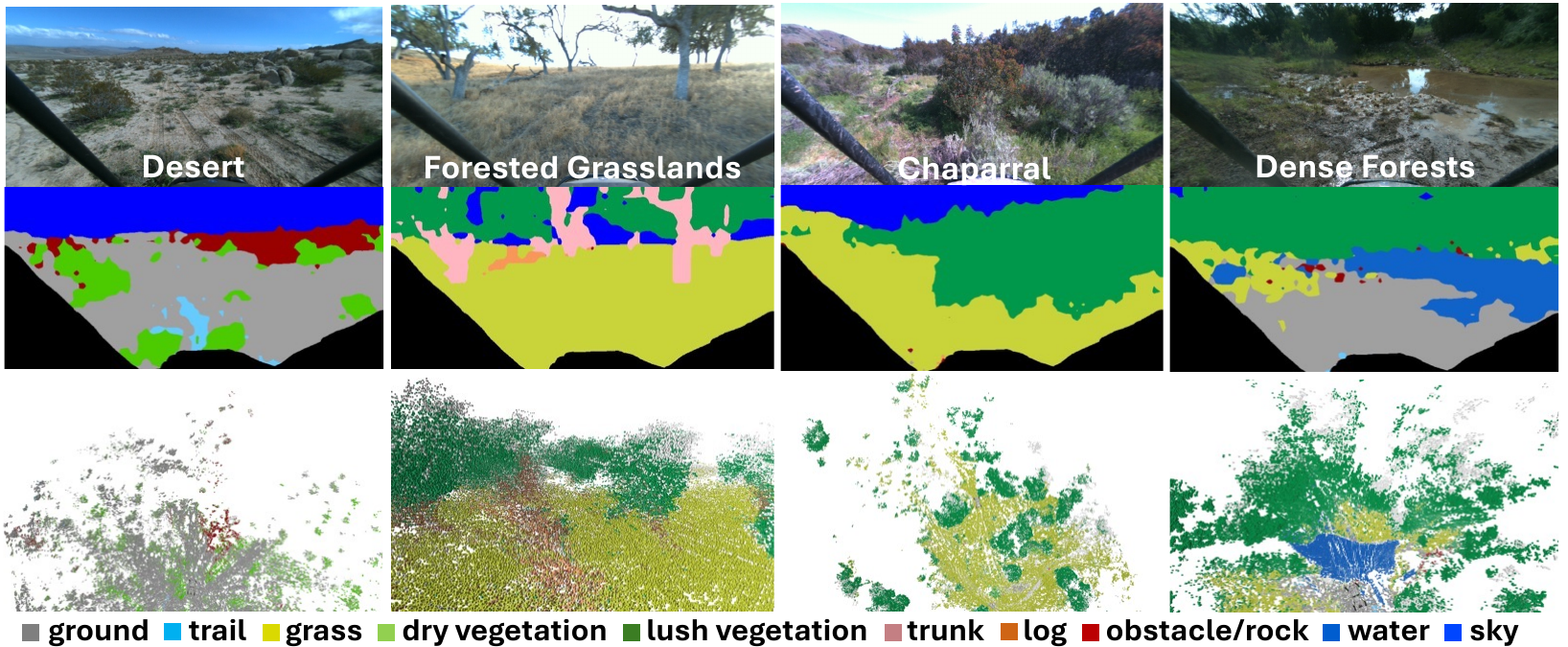}
    \caption{A high-level view of the focus of this work with samples from four different biomes. Each biome sample has an image, a 2D semantic prediction, and a corresponding semantic voxel map colorized by class.}
    \label{fig:sem_map_intro}
\end{figure*}

    
    
    




Resilient, high-speed autonomous off-road driving, both on- and off-trail, has many applications from search and rescue, mining, science data collection, and defense, to extraplanetary robotic exploration.
These environments pose significant challenges for autonomous driving due to the diversity of vegetation types, occlusions, negative obstacles such as ruts and ditches, lighting conditions, and weather impacts.
A robust, general, and adaptable semantic mapping solution is an enabling technology for autonomous driving in these diverse off-road environments.
Creating such a mapping solution is difficult with many approaches requiring large amounts of detailed ground truth labeling \cite{terrainnet} or relying purely on self-supervision \cite{cmu_learn_on_drive,frey2024roadrunner,jpl_robin_trav,chung2024pixel}.
While these approaches are promising, they rely on large datasets with several thousand samples on the low end.
In contrast, learning semantic understanding in a sparse, few-shot manner enables rapid generalizations to unseen biomes, new weather conditions, and other novel scenarios efficiently.

In addition to semantic understanding, perceiving precise geometric information can aid in downstream traversability estimations when operating in unstructured off-road environments with overhanging vegetation from branches, and negative obstacles such as ditches and ruts or dense vegetation. 
A voxel map provides a precise and stable geometric representation of the world.
Used in conjunction with 2D semantic information, voxel maps can provide accurate 3D semantic information without the need for additional labels.

We hypothesize there exists potential to incorporate only small amounts of human expert information in the form of targeted labels to operate in a variety of conditions and biomes.
In this work, we propose leveraging foundation models, coarse and sparse ground truth semantic labels, high-level semantic classes, and a range-based semantic voxel map fusion to create a generalized 3D semantic map.
Our approach supports high-speed autonomous off-road driving in multiple biomes in a few-shot and zero-shot manner.
Figure \ref{fig:sem_map_intro} highlights the multi-biome semantic mapping presented in this work with a showcase of semantic images and maps for different terrains.
In addition to supporting autonomous navigation, a semantic map provides a platform to adapt quickly to new environments and to develop and build a more granular understanding of the world or to support direct traversability learning in a self-supervised manner demonstrated by the RoadRunner architecture~\cite{frey2024roadrunner,roadrunnerM&M} which uses this semantic map to generate its hindsight ground truth data.
The main contributions of this work are as follows:
\begin{itemize}[noitemsep]
    \item Fine-tuning a pre-trained \ac{vit} on a small, sparse, and coarse multi-biome ground truth dataset ($<$500 images, $<$30\% pixels) for efficient zero-shot generalization to new biomes demonstrated by 52.9 and 55.5 mIoU scores on the Yamaha and Rellis datasets respectively.
    \item Few-shot adaptation of a pre-trained \ac{vit} to novel biomes by adding 50 or less sparse, coarse samples which improved mIoU performance on the Yamaha and Rellis datasets to 66.6 and 67.2 respectively.
    \item A novel range-based semantic fusion within a voxel map to enable rapid voxel updates while maintaining temporal stability of 3D semantic maps.
\end{itemize}

\section{Related Works}

\subsection{Off-Road Semantic Segmentation}
\begin{figure*}[t]
    \centering
    \begin{subfigure}[t]{0.24\textwidth}
        \centering \includegraphics[width=\textwidth]{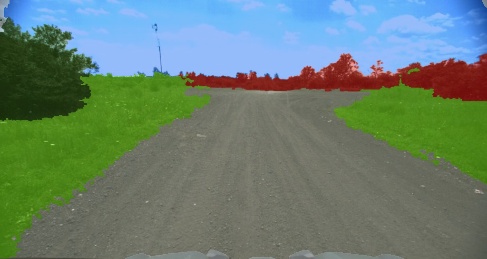}
        \caption{}
    \end{subfigure}
    \begin{subfigure}[t]{0.24\textwidth}
        \centering \includegraphics[width=\textwidth]{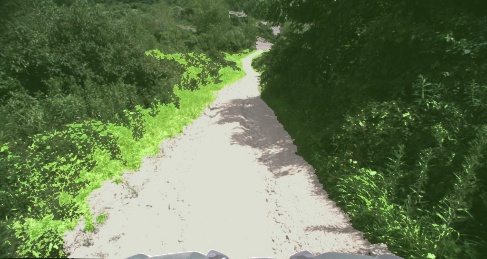}
        \caption{}
    \end{subfigure}
    \begin{subfigure}[t]{0.24\textwidth}
        \centering \includegraphics[width=\textwidth]{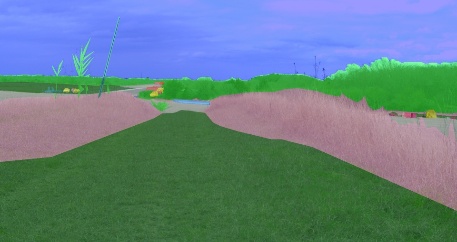}
        \caption{}
    \end{subfigure}
    \begin{subfigure}[t]{0.24\textwidth}
        \centering \includegraphics[width=\textwidth]{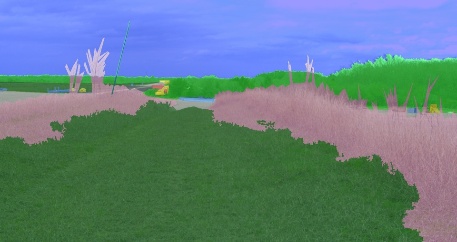}
        \caption{}
    \end{subfigure}
    \caption{Samples from the Yamaha and Rellis datasets with inaccurate and noisy ground truth labels. (a) Treeline labeled as obstacle, (b) noisy grass and trail labels intermixed with high vegetation in the trees, (c \& d) two samples of a similar scene with different precision and class segmentation in the bushes.}
    \label{fig:poor_yamaha}
\end{figure*}

There has been significant work in semantic segmentation especially with the release of popular segmentation datasets like~\cite{Cordts2016Cityscapes}.
For off-road semantic segmentation, some works have adapted architectures for 2D semantic segmentation~\cite{ga-nav,semantic_transfer_seg,offseg} as well as attempts at building multi-modal approaches on top of them~\cite{season_invariant_SS,terrainnet}.
While many approaches have demonstrated high \ac{miou} scores, they have been limited to single environments using datasets with dense labels such as Yamaha~\cite{cmu_semantic_mapping} and Rellis~\cite{jiang2020rellis3d}.
Generating thousands of densely labeled images can incur a high cost, be prone to labeling errors (Figure \ref{fig:poor_yamaha}), especially at the object boundaries, and limit dataset combinations due to the differences in class ontology. 
As a consequence, techniques like this can be difficult to efficiently adapt to a new environment.
Some approaches to dataset creation have generated multiple labels per image to reduce noise~\cite{ai4mars} or attempted to detect poor training samples for re-annotation~\cite{BECKTOR2022100411}.
Other approaches have attempted to forgo any semantic labels and attempt to learn in a self-supervised manner based on trajectories from either offline \cite{jpl_robin_trav} or online via near-range LiDAR geometry \cite{cmu_learn_on_drive}.
However, while these approaches have demonstrated performance on separating 2-3 classes, they are still experimental and limited in their ability to identify the semantic variation and complexity encountered in multi-kilometer high-speed autonomous off-road driving targeted in this work.
Recent ideas regarding the importance of dataset quality in the area of data-centric \ac{ai}~\cite{data_centric_ai} and work that demonstrated the potential of using small, sparse datasets when traversing to new regions of extraplanetary surfaces~\cite{atha_spoc} provide a basis for this work.

\subsection{Semantic Voxel Mapping}
A common semantic aggregation pipeline is the use of a 2.5D layered grid map such as in~\cite{cmu_semantic_mapping}. While these representations are effective for many environmental terrains, they do not capture certain geometry, most notably overhangs. 
Voxel maps have been utilized in robotic mapping due to their ability to aggregate and fuse complex 3D features~\cite{voxel1,voxel2}. Voxel maps have been used in applications where mapping ranges or voxel resolutions are limited~\cite{voxblox} since traditionally voxel maps have been memory intensive and slow. With recent advances in \ac{gpu} technology, voxel mapping can now be run in real-time for mobile robotics as demonstrated by GVOM~\cite{overbye2021gvom}. However, GVOM only handles geometric aggregation and does not handle the fusion of semantic values over time. 

There has been recent work on semantic fusion within a voxel map based on Bayesian updates~\cite{bayesian_mapping}. This technique relies on Bayesian probabilities being reported by the semantic segmentation, which presents a computation and latency challenge limiting its feasibility for low latency high-speed driving. Other approaches based on voting~\cite{voxel_voting1} and averaging~\cite{voxel_average2} have been applied in confined indoor environments. These methods in an off-road environment can be skewed by different speeds and slow to react to pop-up hazards.

\section{Method}




\begin{figure*}[t]
    \centering
    \includegraphics[width=\textwidth]{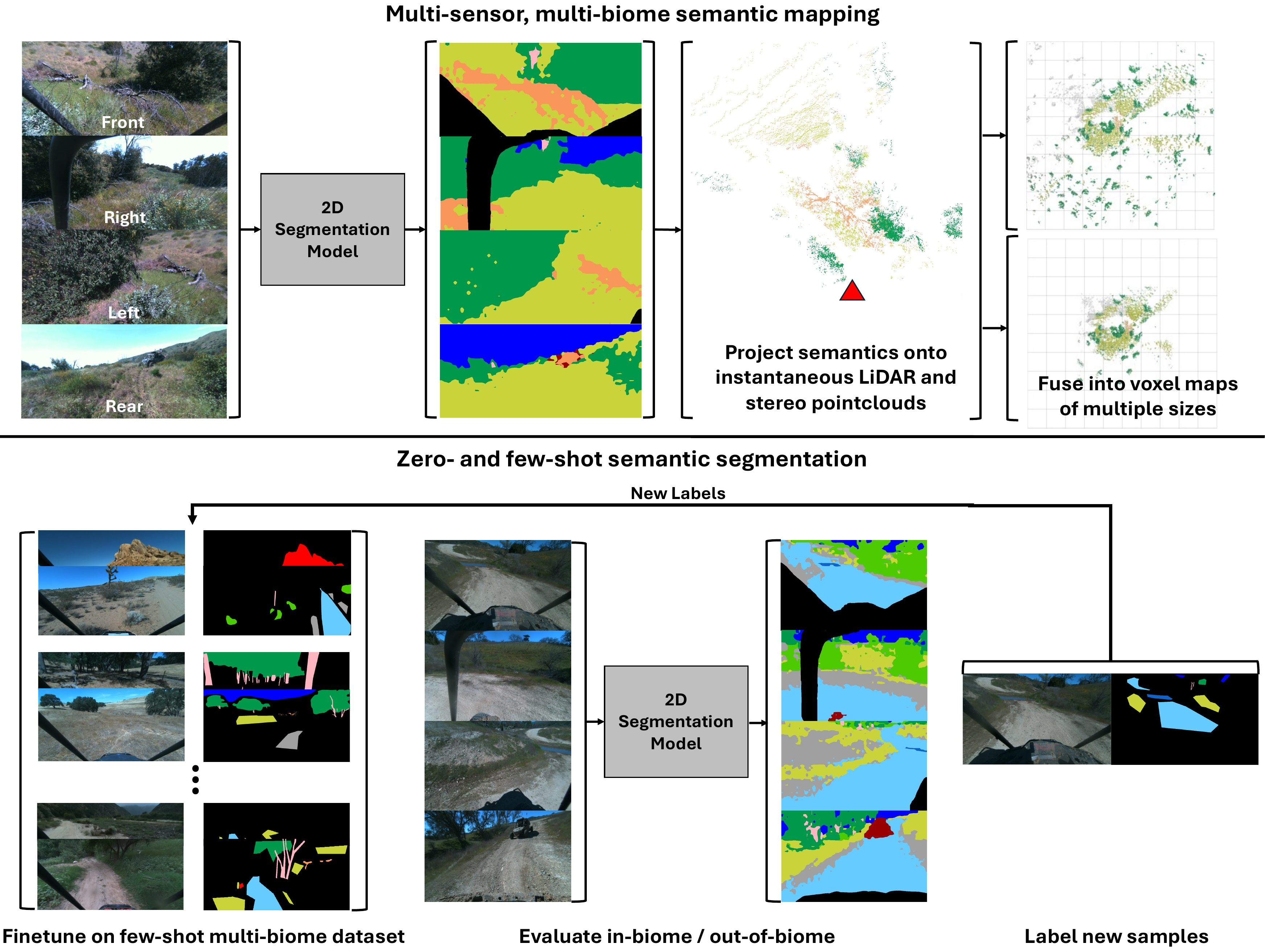}
    \caption{(top) Our architecture for semantic voxel mapping using multiple cameras and LiDARs. (bottom) Our process to create a small, diverse multi-biome dataset of coarse, sparse ground truth labels. This architecture and dataset approach enables zero- and few-shot multi-sensor, multi-biome 3D mapping at multiple sizes and ranges.}
    \label{fig:sem_map_arch}
\end{figure*}

\subsection{Semantic Mapping Architecture}
\label{sec:arch}

To accomplish high-speed, multi-biome off-road semantic mapping for biomes both with and without existing data, we propose a three-part architecture to achieve zero- and few-shot 3D semantic mapping that is robust to complex, unstructured geometry.
Firstly, we train a model to infer 2D semantic segmentation using few-shot coarse, sparse semantic labels to generalize across biomes.
Additional labels can be added to the dataset as new biomes, conditions, and challenging scenarios are experienced to improve the generalization of the model.
Secondly, we utilize LiDAR and stereo 3D information in order to project the semantic segmentation classes into 3D space.
Thirdly, we fuse the 3D semantic information with a range-based metric into a voxel map to produce an aggregate 3D semantic map from multiple cameras and LiDARs over time.
Figure \ref{fig:sem_map_arch} highlights this architecture. The top row shows the process of building the multi-biome semantic segmentation dataset and the bottom row shows the three-part architecture to produce a 3D semantic map.

\subsubsection{Image Semantic Segmentation}
\label{sec:image_semantics}
We perform image-based semantic segmentation to classify every pixel within an image to a semantic class.
We employ the \ac{vit} S-16 encoder with the Segmenter architecture \cite{segmenter} due to demonstrated robustness of \acp{vit} to domain-shifts, severe occlusions, and spatial permutations~\cite{vit_properties}.
The \ac{vit} encoder is pre-trained on ImageNet-21k with strong data augmentation and regularization \cite{how_to_train_vit} and provided by mmsegmentation \cite{mmseg2020}.
We choose these semantic classes: ground, trail, grass, dry vegetation, lush vegetation, trunk, log, obstacle/rock, water, and sky based on easy interpretation and being vehicle agnostic.
However, new classes can easily be added due to the few-shot capabilities of our method.
We fine-tune the model using a weighted cross-entropy loss due to the natural imbalance of classes in off-road environments. 
For data augmentation, we applied resizing, random cropping,  and random flipping.

\subsubsection{Pointcloud Projection}
Next, we backproject each point in the LiDAR pointcloud into the image space taking into account the vehicle pose~\cite{fakoorian2022rose,nissov2024roamer}, the LiDar-camera extrinsic and the camera intrinsic calibration parameters.
Each point receives the semantic class estimate from the previous step, leading to a semantic pointcloud.
We introduce a special case for water segmentation.
We do not use LiDAR pointclouds for mapping water, due to the reflection issue.
Instead, water is projected onto a stereo pointcloud.
Nevertheless, stereo can also have issues with holes and reflections \cite{water_reflection}. 
Therefore, a plane fit of the water surface is computed using all the water points to find its surface height.

\subsubsection{Voxel Map Fusion}
\label{sec:sem_fusion}
To aggregate semantics into the voxel map, we select the semantic values of a 3D point observed closest to the vehicle using:
\begin{equation}
\label{eq:nearest_semantics}
    \begin{split}
    C_{i_{vox},j_{cls}} = \begin{cases} 
      C_{i_{vox},j_{cls}} & R_{i_{vox}} \leq R'_{i_{vox}} \ \mathrm{or} \  C'_{i_{vox},j_{cls}} = \mathrm{NaN} \\
      C'_{i_{vox},j_{cls}} & \mathrm{otherwise}
   \end{cases} \\
   \forall \quad 1 \leq i_{vox} \leq N_{vox} \quad \textrm{and} \quad 1 \leq j_{cls} \leq N_{cls}
   \end{split}
\end{equation}
where $C$ is the existing confidence of a voxel for a specific class, $C'$ is a confidence of a potential incoming semantic measurement, $R$ is the range at which the semantic measurement was captured, $R'$ is the range to a potential new semantic measurement, $i_{vox}$ is the index of a voxel within the voxel map, $j_{cls}$ is the index of a class in the set of all class, $N_{vox}$ is the number of voxels, and $N_{cls}$ is the number of semantic classes.

There are three primary motivations for this fusion approach: (1) rapid reaction to pop-up hazards, (2) ability to correct for bleeding issues that are prevalent when using image-to-LiDAR projection, and (3) map stability especially when vehicle drives in reverse.
Firstly, pop-up hazards are common in off-road environments primarily due to heavy occlusion such as rocks behind bushes or logs hidden in grass.
This approach provides a rapid map response to these hazards, enabling higher-speed driving.
Secondly, image-to-LiDAR projection typically contains more bleeding at range and prioritizing closer measurements minimizes the impact over time.
Thirdly, the map will not degrade when regions are later viewed further away.
This limits bleeding and occlusion impacts in previously well perceived regions during vehicle reverses and double backs.

\subsection{Dataset for Few-Shot Multi-Biome Semantic Segmentation}
\label{sec:dataset_sem}
The key strength of this method is that it requires a minimal number of labels.
A model such as Segmenter~\cite{segmenter} with a pre-trained \ac{vit}, reduces the data complexity required to capture an environment.
Another strength of this approach is using sparse and coarse labeling to create high-precision ground truth data.
Most dense off-road datasets contain boundary noise and class errors like the samples in Figure \ref{fig:poor_yamaha}.
Based on recent proposals in data-centric \ac{ai}~\cite{data_centric_ai}, we believe minimizing noise within the labels will minimize the data required to train a model.

\subsubsection{Data Collection and Selection}
In order to generate a diverse dataset for this work, a subset of images are chosen from a variety of terrains and daytime weather conditions, excluding snow conditions, that are representative of an off-road driving scenario.
These terrains included grasslands/western forests, chaparral, and desert biomes.  This data is collected with our Polaris RZR vehicle, which is detailed in Section \ref{sec:vehicle}.
We select initial data for labeling based on human identification of images with challenging lighting, minority classes, and complex scenarios from both human and autonomous driving.
Additional data is continuously selected based on instances of unplanned human intervention or when the vehicle reverses under autonomous operation. 

\subsubsection{Data Labeling}

The highest priorities for our labeling scheme are to minimize noise and label error and to label an image quickly.
We propose an off-road labeling methodology that enables both fast and low-error labels:
(1) Ensure all labels do not pass a class boundary. It is preferable to have smaller, precise labels.
(2) Label only those pixels where one is confident they are the class being labeled. Objects with only a few visible pixels such as those far away or highly occluded can be ignored.
(3) Focus labeling on minority classes, challenging scenarios and their surrounding regions. For example, labeling the grass or ground around a rock while mostly ignoring large open ground or grass fields.
The resulting labels will be both sparse and coarse.
Figure \ref{fig:sem_map_arch} provides a few example labels in the bottom right and bottom left regions.
This approach enables rapid labeling, and therefore, rapid model adaptation for new environments and has been used by the team to train improved models same day in previously unseen biomes.

\subsection{Vehicle Sensors and Compute}
\label{sec:vehicle}
The vehicle platform for this work contains a sensor setup with three Velodyne VLP-32 LiDARs, two on the front and one in the rear, and four Multisense S27 cameras with three on the front and one in the rear.
The vehicle has additional sensors, which were not utilized in the context of this work.
The vehicle compute consists of a Threadripper 3990x CPU (64 Core 2.9/4.3 GHz), 256 GB RAM, and 4 $\times$ GeForce RTX 3080 GPUs.
Due to the requirements of other autonomy stack components, the semantic segmentation and voxel mapping processes are each allocated one GPU.
The three front cameras and lidars operate at 10hz.
The back camera operates at 2hz. 
The voxel mapping processes fuse the 10hz semantic pointcloud information and produces an aggregated map at 5hz.

\section{Experiments and Results}


\subsection{Datasets for Image Semantics Evaluation}
\label{sec:eval_dataset}

\begin{table}
\caption{The class remapping amongst datasets to maintain consistency with our existing multi-biome dataset.}
\label{tab:remapping}
\begin{center}
\begin{adjustbox}{width=1\columnwidth}
\begin{tabular}{p{0.22\linewidth}|p{0.28\linewidth}|p{0.22\linewidth}| p{0.28\linewidth}}
\hline
\textbf{New Class} & \textbf{Ours} & \textbf{Yamaha} & \textbf{Rellis} \\
\hline
ground & ground, trail & trail, rough trail & asphalt, mud, concrete \\
\hline
grass & grass & grass, non traversable low veg. & grass \\
\hline
vegetation & trunk, dry veg., lush veg. & high veg. & tree, bush \\
\hline
obstacle & obstacle, log & obstacle & log, container, vehicle, pole, barrier, rubble, fence, person, building \\
\hline
water & water & puddle & water, puddle \\
\hline
sky & sky & sky & sky \\
\hline

\end{tabular}
\end{adjustbox}
\end{center}
\end{table}

\begin{table*}
\caption{The composition of our dataset broken down by a number of images, percentages of all pixels that are labeled, and the percentage of labeled pixels that are a specific class. This encompasses a multi-biome dataset and additional re-labeled samples from the Yamaha and Rellis datasets.}
\label{tab:dataset}
\begin{center}
\begin{adjustbox}{width=1\textwidth}

\begin{tabular}{p{0.27\linewidth}cccccccc}
\hline
Biome & Images & Px labeled & Ground & Grass & Veg. & Obstacle & Water & Sky \\
\hline
Mojave Desert & 174 & 24.7 & 55.8 & 0.0 & 7.5 & 8.4 & 0.0 & 28.3 \\
Paso Robles Grassland & 197 & 30.3 & 20.5 & 42.4 & 25.6 & 1.7 & 0.0 & 9.7 \\
San Diego Grassland & 30 & 21.6 & 20.2 & 32.4 & 44.3 & 0.3 & 0.9 & 1.9 \\
S. California Chaparral & 84 & 14.7 & 43.6 & 5.1 & 12.0 & 6.4 & 30.4 & 2.6 \\
Misc. Online & 11 & 36.9 & 9.2 & 21.4 & 24.4 & 7.2 & 8.4 & 29.4 \\
\hline
Total & 496 & 25.6 & 34.5 & 23.7 & 19.0 & 4.5 & 2.7 & 15.6 \\
\hline
Yamaha & 50 & 20.3 & 39.4 & 21.3 & 20.0 & 13.3 & 1.8 & 4.1 \\
Rellis & 50 & 12.5 & 10.0 & 40.2 & 25.3 & 7.3 & 3.3 & 13.8 \\
\hline

\end{tabular}
\end{adjustbox}
\end{center}
\end{table*}

We evaluate the zero- and few-shot semantic segmentation approach outlined in this work on the open-source off-road datasets Yamaha \cite{season_invariant_SS} and Rellis \cite{jiang2020rellis3d}.
The zero-shot evaluation is performed on test sets from Yamaha and Relllis with a model trained on a multi-biome dataset created following the method in Section \ref{sec:dataset_sem}.
The class ontology of Relllis, Yamaha, and ours are different; therefore, we perform a remapping outlined in Table \ref{tab:remapping}.
We additionally relabeled 50 images from both the Yamaha and Rellis datasets according to our labeling methodology.
We selected, via human inspection, the 50 image samples from Yamaha and Rellis separately for relabeling from problematic zero-shot predictions from the full set of training samples.
Quantitative metrics such as selecting the worst \ac{miou} scores were rejected for two reasons.
Firstly, noise and errors in the existing ground truth skewed the sample selection.
Secondly, newly collected data for inclusion into the dataset will not have ground truth data. 
We minimize the need for human inspection by utilizing instances of poor autonomous operation such as human interventions to narrow the search space.
This approach is not possible on the Yamaha and Rellis datasets, and therefore human inspection is used as a proxy.
After relabeling, the composition of the multi-biome dataset including the Yamaha and Rellis samples in Table \ref{tab:dataset}.
We trained models on different combinations of this relabeled data and our multi-biome dataset and evaluated them on their respective test sets, consistent with the zero-shot evaluation.

To evaluate model performance on a test set, we used the original ground truth labels of Yamaha and Rellis.
While this approach is sub-optimal, it prevented bias from using our ground truth labels.
Unfortunately, neither Yamaha nor Rellis has a specific test set that has been made open-source.
Therefore, for the Yamaha dataset, we used the validation set with the bottom 32 rows cropped to remove the vehicle.
It only contains a few pixels of labeled water from a single image.
To address this limitation, we selected and added five additional samples from the training set to the test set.
These samples were selected based on having the most accurate ground truth labels. 
The total test set contained 150 images.
For Rellis, there was no separation of the data.
Therefore, 101 images were randomly sampled and resized to 960$\times$600 to generate a test set.

\subsection{Image Semantics Evaluation}

We trained models on different combinations of the data outlined in Section \ref{sec:eval_dataset} for comparison.
The results of this evaluation are in Table \ref{tab:sem_quant}.
We observe strong initial zero-shot performance on the dominant classes and weaker performance on minority classes.
The performance of the obstacle class on Yamaha was expected since it contains more man-made obstacles such as street signs, metal fences, and buildings than exist in our dataset.
Better zero-shot performance on water would be desired but is not unexpected as the water samples in Yamaha and Rellis are primarily puddles, whereas the water in our dataset is often larger water features such as creeks, streams, and flooded regions.
Fortunately, these differences provided a means to demonstrate few-shot dataset adaptation to capture the features of these environments.

Few-shot dataset adaptation is evaluated by performing tests with 10, 25, and 50 samples added to our existing dataset.
These samples were split 80\% into the train set and 20\% into the validation set.
We observe \ac{miou} increases on Yamaha and Rellis with the inclusion of 10, 25, and 50 samples.
Most classes saw some \ac{iou} gains with the most improvement from the obstacle and water \acp{iou}.
We also observe that while using only the 50 labels from Yamaha and Rellis was able to outperform the zero-shot, a model trained on the existing multi-biome dataset and additional in-biome labels performed the best.
This supports our hypothesis that models can generalize across biomes using small, targeted datasets from many biomes.

\begin{table}
\caption{The \ac{iou} and \ac{miou} scores for different dataset approaches for training zero- and few-shot multi-biome capable models. A pixel is considered to be predicted a class if a model output has a confidence score of 0.5 or greater.
}
\label{tab:sem_quant}
\begin{center}
\begin{adjustbox}{width=1\columnwidth}
\begin{tabular}{lp{.16\linewidth}p{0.16\linewidth}|cccccc|c}
\hline
Dataset & \centering Out-of-Biome Samples & \centering In-Biome Samples & Ground & Grass & Veg. & Obstacle & Water & Sky & mIoU\\
\hline
 & \centering 496  & \centering 0    & 70.3   & 40.3   & 73.7   & 33.0   & 22.0   & 78.4   & 52.9 \\
 & \centering 496  & \centering 10   & 78.2   & 55.4   & 71.7   & 59.7   & 30.6   & 77.2   & 62.2 \\
Yamaha & \centering 496  & \centering 25   & 78.5   & 54.7   & 72.3   & 65.2   & 21.7   & 77.9   & 61.7 \\
 & \centering 496  & \centering 50   & 81.5   & 63.3   & 74.4   & 68.3   & 32.3   & 79.2   & 66.6 \\
 & \centering 0   & \centering 50   & 79.8   & 63.7   & 72.3   & 61.9   & 21.8   & 64.5   & 60.7 \\
\hline
 & \centering 496  & \centering 0    & 58.3   & 69.5   & 56.3   & 53.8   & 0.0   & 95.3   & 55.5 \\
 & \centering 496  & \centering 10   & 57.1   & 73.2   & 75.5   & 59.9   & 17.5   & 95.3   & 63.0 \\
Rellis & \centering 496  & \centering 25   & 52.1   & 76.5   & 78.6   & 62.4   & 38.2   & 95.3   & 67.2 \\
 & \centering 496  & \centering 50   & 53.1   & 78.4   & 78.7   & 57.5   & 37.7   & 94.9   & 66.7 \\
 & \centering 0   & \centering 50   & 32.8   & 71.5   & 66.0   & 62.1   & 35.2   & 92.5   & 60.0 \\
\hline

\end{tabular}
\end{adjustbox}
\end{center}
\end{table}

\begin{figure*}
    \centering
    \includegraphics[width=\textwidth]{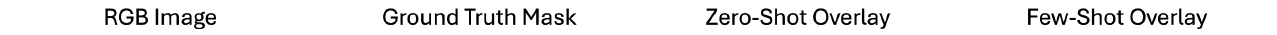}
    \includegraphics[width=0.235\textwidth]{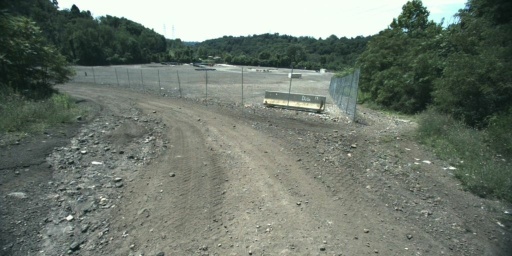}
    \includegraphics[width=0.235\textwidth]{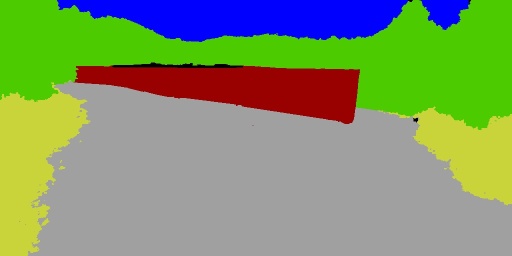}
    \includegraphics[width=0.235\textwidth]{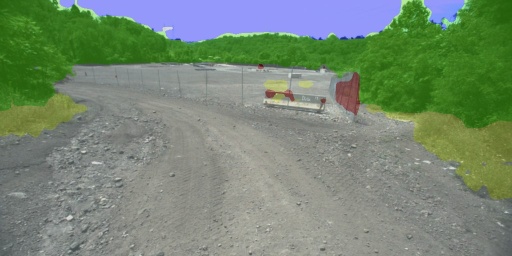}
    \includegraphics[width=0.235\textwidth]{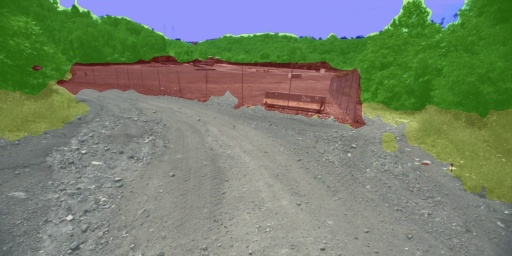}

    \includegraphics[width=0.235\textwidth]{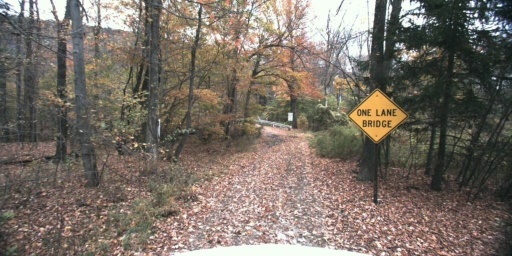}
    \includegraphics[width=0.235\textwidth]{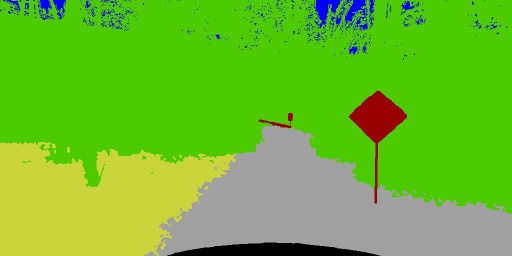}
    \includegraphics[width=0.235\textwidth]{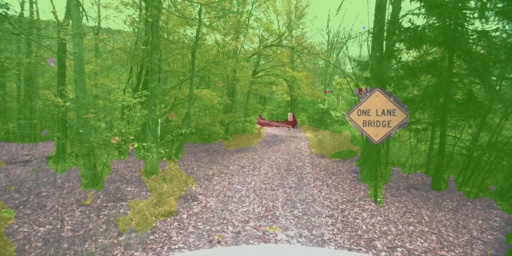}
    \includegraphics[width=0.235\textwidth]{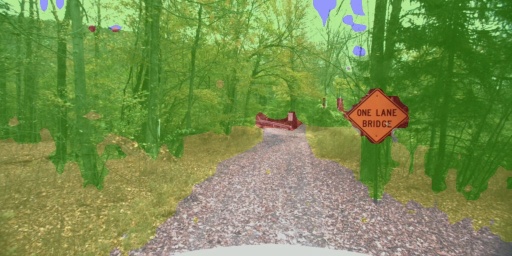}

    \includegraphics[width=0.235\textwidth]{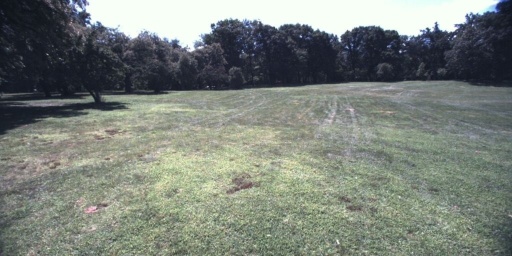}
    \includegraphics[width=0.235\textwidth]{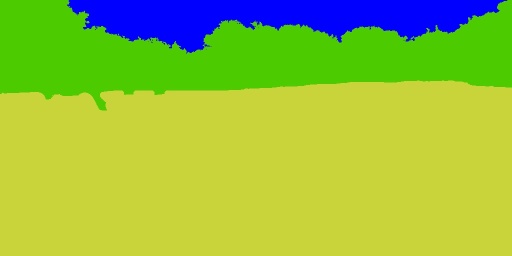}
    \includegraphics[width=0.235\textwidth]{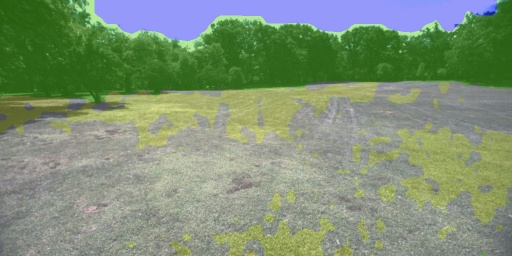}
    \includegraphics[width=0.235\textwidth]{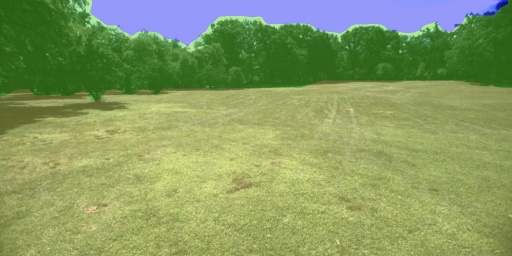}

    \includegraphics[width=0.235\textwidth]{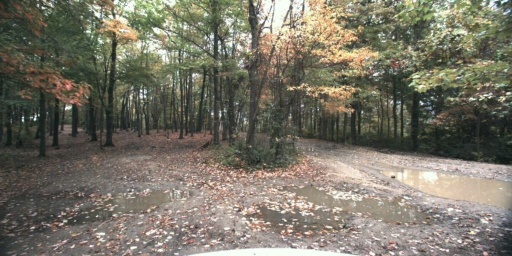}
    \includegraphics[width=0.235\textwidth]{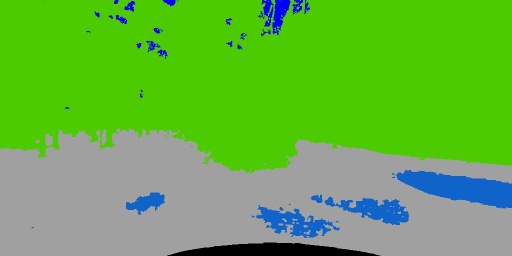}
    \includegraphics[width=0.235\textwidth]{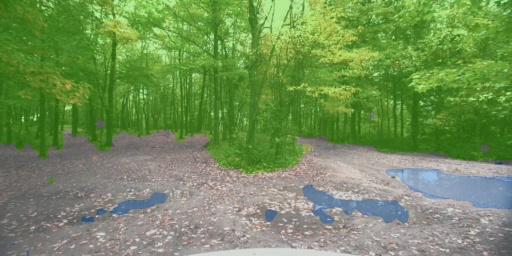}
    \includegraphics[width=0.235\textwidth]{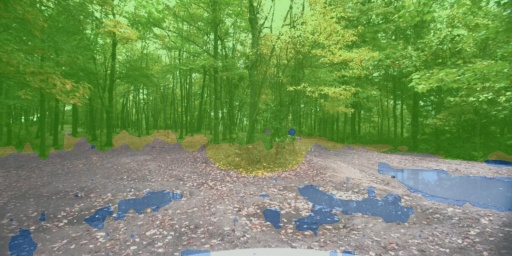}

    \includegraphics[width=0.235\textwidth]{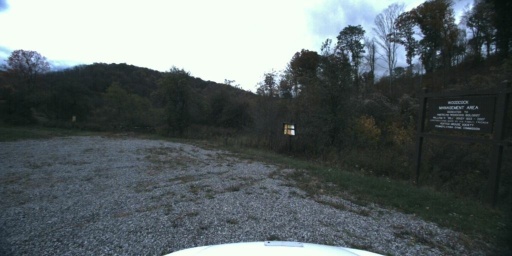}
    \includegraphics[width=0.235\textwidth]{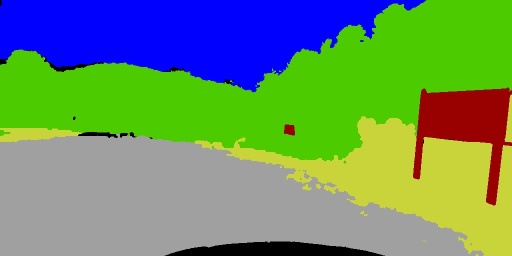}
    \includegraphics[width=0.235\textwidth]{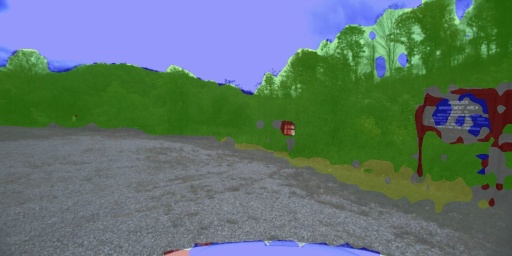}
    \includegraphics[width=0.235\textwidth]{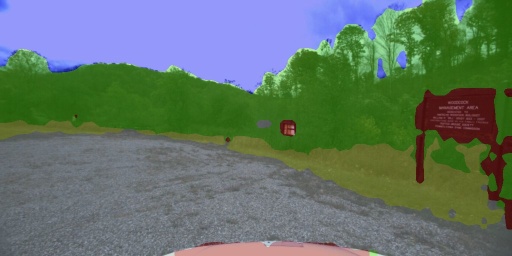}

    \includegraphics[width=0.235\textwidth]{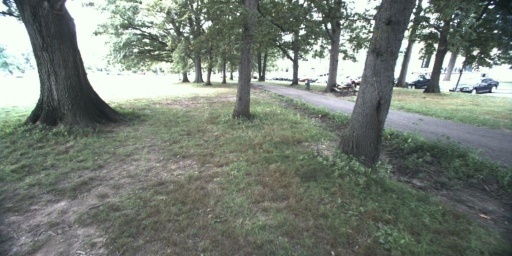}
    \includegraphics[width=0.235\textwidth]{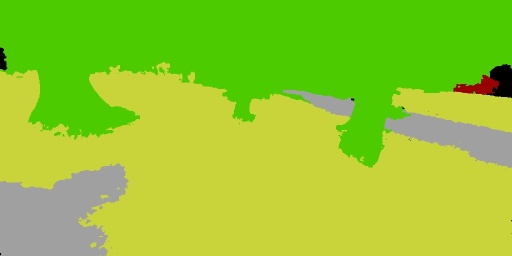}
    \includegraphics[width=0.235\textwidth]{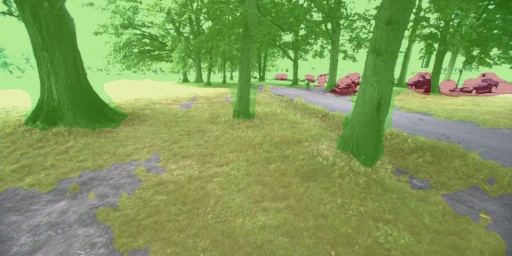}
    \includegraphics[width=0.235\textwidth]{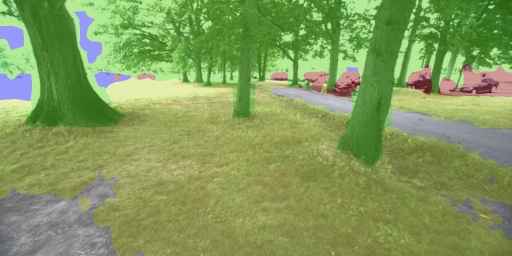}

    \includegraphics[width=0.235\textwidth]{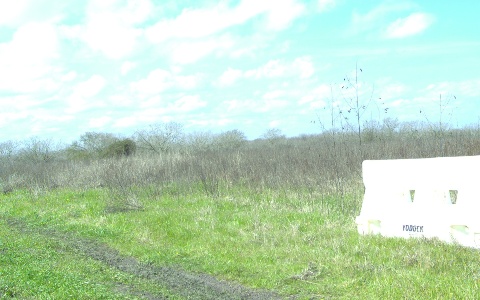}
    \includegraphics[width=0.235\textwidth]{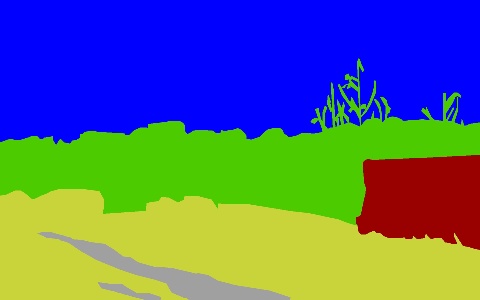}
    \includegraphics[width=0.235\textwidth]{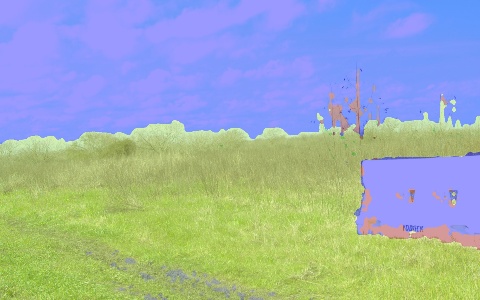}
    \includegraphics[width=0.235\textwidth]{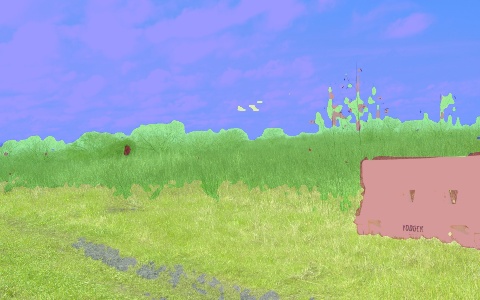}

    \includegraphics[width=0.235\textwidth]{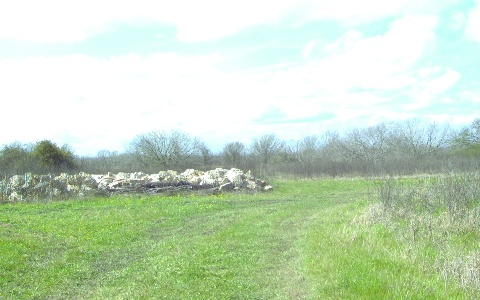}
    \includegraphics[width=0.235\textwidth]{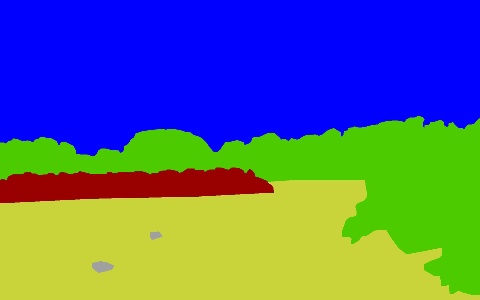}
    \includegraphics[width=0.235\textwidth]{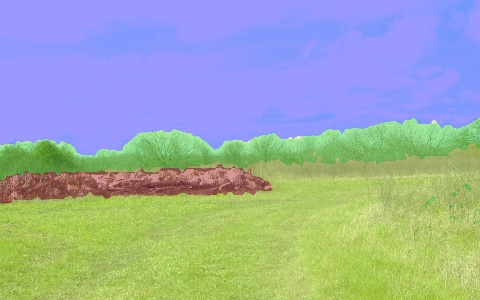}
    \includegraphics[width=0.235\textwidth]{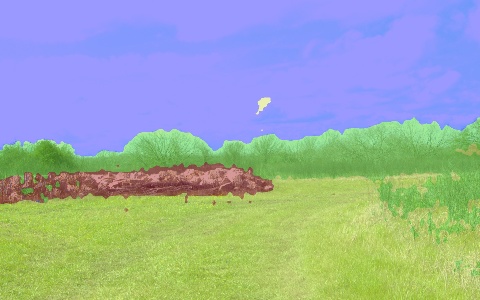}

    \includegraphics[width=0.235\textwidth]{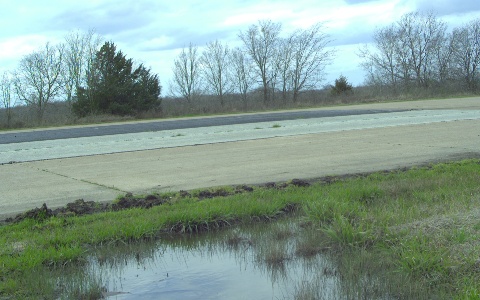}
    \includegraphics[width=0.235\textwidth]{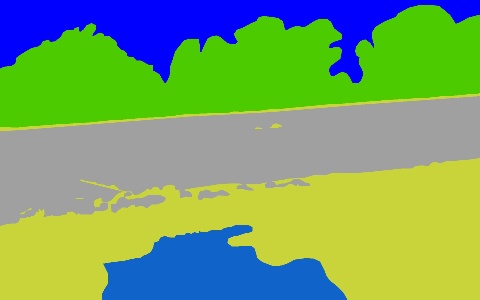}
    \includegraphics[width=0.235\textwidth]{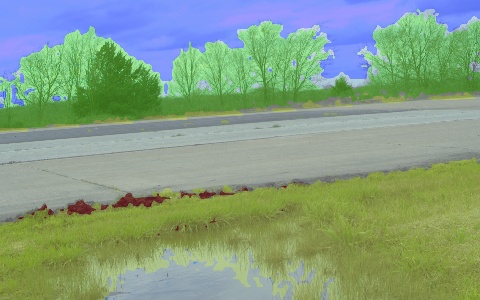}
    \includegraphics[width=0.235\textwidth]{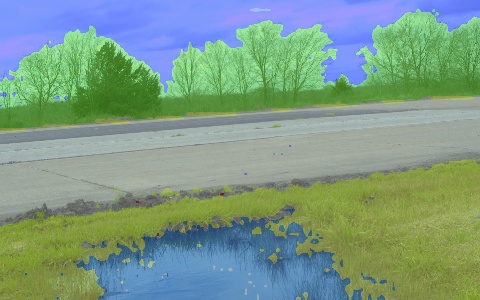}

    \centering \includegraphics[width=0.6\textwidth]{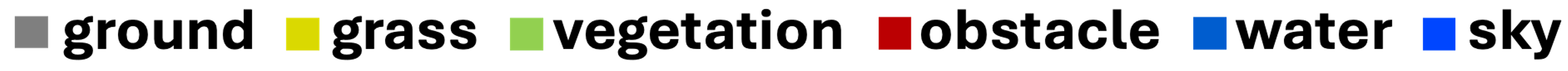}
    
    \caption{Qualitative samples from the Yamaha(top 6 rows) and Rellis datasets (bottom 4 rows) with the RGB image, ground-truth mask, zero-shot prediction overlay, and few-shot prediction overlay from the top \ac{miou} model. Zero-shot performance predicts most of the region correctly. Some classes such as water and obstacle are improved with additional in-biome samples in the training dataset.}
    \label{fig:qual_yamaha_rellis}
\end{figure*}

\begin{figure*}[t]
    \centering
    \includegraphics[width=\textwidth]{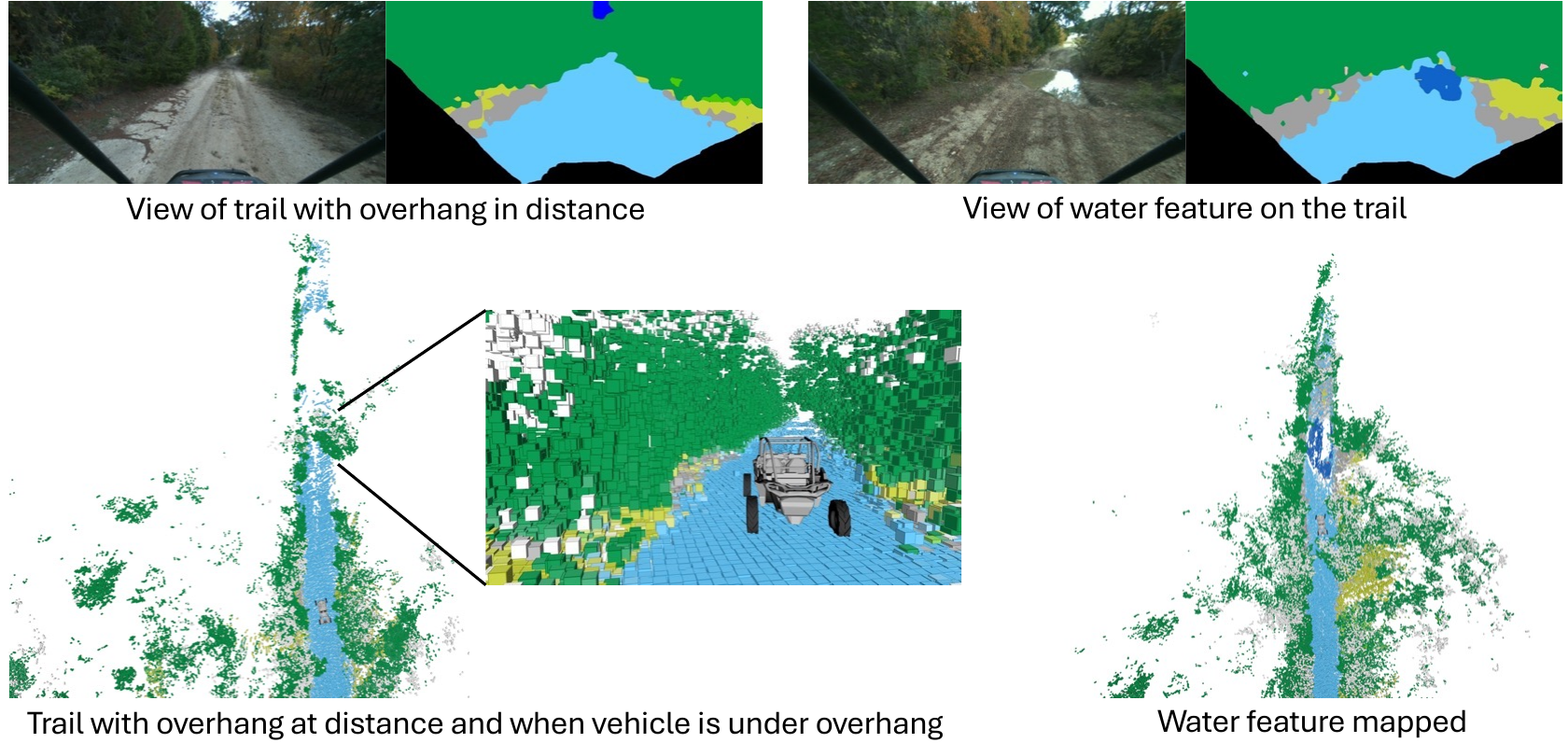}
    \caption{Semantic mapping through time in an overhang environment with a water hazard (dark blue).}
    \label{fig:qual_racer_overhang}
\end{figure*}

The quantitative results presented above are supported by qualitative samples from the two test datasets in Figure \ref{fig:qual_yamaha_rellis}.
For the Yamaha dataset, we observed that the zero-shot model is able to successfully identify most vegetation, some water features, and some hazards such as cars and a couple of signs.
A model trained with the additional 50 in-biome samples was able to detect previously unseen signs and a metal fence.
There is confusion between the ground and grass classes in the zero-shot model, which is resolved with the the 50 additional samples.
For the Rellis dataset, we observe similar zero-shot capabilities with rock piles, grass, and ground being successfully segmented.
A model trained with the additional 25 in-biome samples was able to segment the water and some previously not labeled obstacles.

\subsection{Voxel Semantic Fusion}

A semantic voxel map provides advantages for off-road driving due to its precision and ability to represent complex 3D geometry such as overhanging tree branches and canopies. 
Traversability risks due to overhangs are difficult to manually ground truth because their danger is dependent upon where the ground plane is.
A voxel map allows for precise geometric mapping of semantic information, enabling informed decisions on overhangs and their risks.
In Figure \ref{fig:qual_racer_overhang}, we observe a sample of an overhang during autonomous driving along a trail.
The voxel map captures the overhang geometry precisely and the vehicle is able to drive along the trail. 
The second aspect observed in Figure \ref{fig:qual_racer_overhang}, is the water feature.
Initially, there is a hole in the map where the water is located because the LiDAR points are reflected and not segmented within the image.
Once detected semantically, the water is successfully mapped via a plane fit of the stereo data described in Section \ref{sec:arch}.
This highlights the capability to fuse multiple modalities and sensors into a single map.

\subsubsection{Rapid fusion updates with stability}
\begin{figure*}[t]
    \centering
    \includegraphics[width=\textwidth]{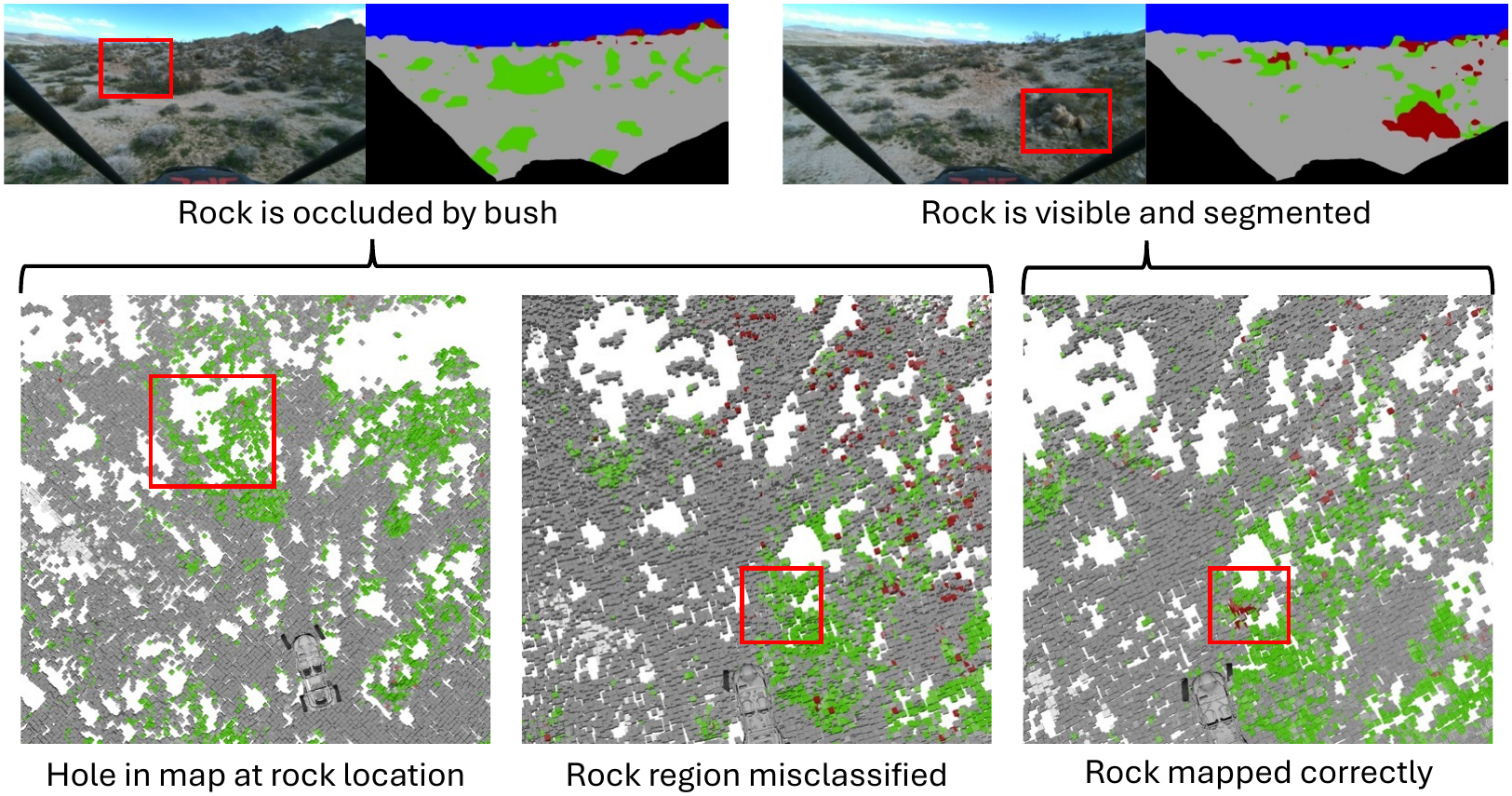}
    \caption{A figure with a sequence of images and voxel maps of an approach to a rock hidden behind a bush. The rock is initially occluded and the vegetation segmentation propagates into the map. Once the rock becomes visible, it is segmented correctly and the map responds accordingly.}
    \label{fig:qual_racer_update}
\end{figure*}

As discussed in Section \ref{sec:arch}, our map should respond quickly to new information while maintaining the stability of regions at close range where our segmentation is confident.
In Figure \ref{fig:qual_racer_update}, we observe an example of this trade-off using our proposed range-based fusion approach.
In this scenario, the vehicle was driving through a lightly vegetated desert environment and was approaching a rock that was occluded by a bush.
Initially, the image segmentation predicted this region as dry vegetation.
Due to the sparsity of the bush, there were LiDAR returns in the region of the rock which were aggregated into the voxel map.
As the vehicle drove forward, the image view changed and the rock became visible.
The image segmentation predicts the rock as obstacle, and the map is updated immediately because this prediction occurred at a closer range.
Furthermore, if the vehicle were to reverse out of the situation, the rock would become occluded again.
The obstacle prediction would persist in the map regardless of how many future dry vegetation predictions were to be added to the voxel map.
This poses an advantage over other fusion approaches such as the Bayesian updates.
A Bayesian update could require multiple predictions to update a voxel from dry vegetation to obstacle, increasing the risk of collision and could erode the obstacle prediction after a reverse causing potential repeated wrong maneuvers.


\subsubsection{Key Challenges}
This approach has a risk for semantic bleeding due to parallax, inaccurate calibrations, inaccurate semantic boundaries, and other impacts.
Our proposed fusion aided in updating bleeding upon approach but could cause issues in the map at longer ranges.
A second challenge was the ability to perform useful quantitative evaluations of the semantic voxel map performance.
The evaluation was performed by human experts rating semantic map performance on a diverse set of representative scenarios from field experiment data.
To develop a more robust evaluation method without a human-in-the-loop, work is ongoing to build hindsight voxel maps for ground truth.
However, these hindsight maps will have a bias since they have to be generated with some existing mapping approach.

\subsection{Multi-Modal Hindsight Ground Truth Generation}
Our mapping approach can be extended to aid in training more complex, multi-modal models.
While a detailed review is outside of the scope of this paper, a more detailed analysis is in our previous work on Road Runner, a multi-modal self-supervised model to predict \ac{bev} traversability maps~\cite{frey2024roadrunner,roadrunnerM&M}.
Based on these works, the potential exists to drive autonomously in a new biome with our zero-shot image segmentation and mapping, efficiently update our image segmentation dataset and retrain, build hindsight ground truth traversability maps of the new biome, and fine-tune a more complex model such as Road Runner for efficient adaptation to a new biome.

\section{Conclusion and Future Work}

In this work, we demonstrate a method for zero- and few-shot multi-biome semantic segmentation utilizing sparse, coarse ground truth labels with $<$500 total images with $<$30\% of their pixels labeled.
An evaluation on the Yamaha and the Rellis datasets demonstrates the ability for zero-shot performance and to adapt our dataset with few-shot in-biome samples to train improved models.
Furthermore, a novel semantic voxel fusion method based on range is proposed, for rapid map response and stability of confident regions.
This method is utilized in multiple biomes and can provide hindsight ground truth data for training multi-modal self-supervised models.
Future work is ongoing to expand the number of biomes within the dataset, improve the evaluation of the semantic voxel mapping via both a hindsight ground truth voxel map framework and a photo-realistic simulation, and to extend the semantic segmentation to more detailed sub-classes unique to different biomes.

\section{Acknowledgements}
The High Performance Computing resources used in this investigation were provided by funding from the JPL Information and Technology Solutions Directorate.

\begin{figure*}
    \centering
    
    \includegraphics[width=0.24\textwidth]{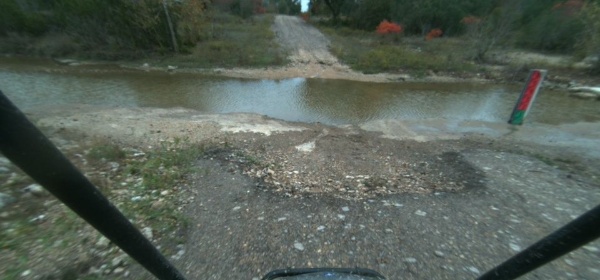} 
    \includegraphics[width=0.24\textwidth]{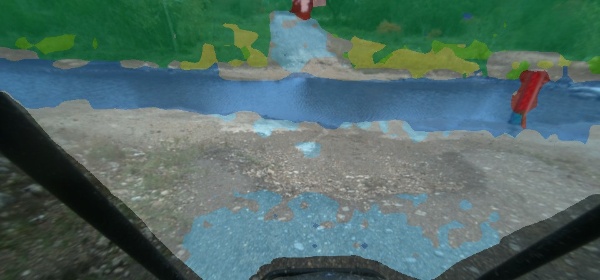}
    \includegraphics[width=0.24\textwidth]{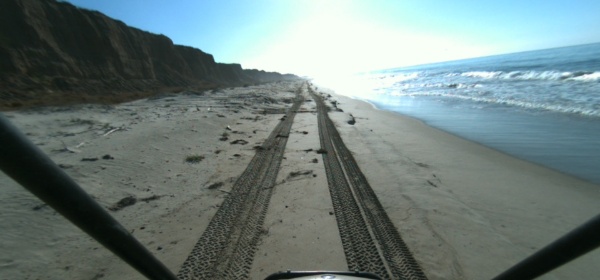} 
    \includegraphics[width=0.24\textwidth]{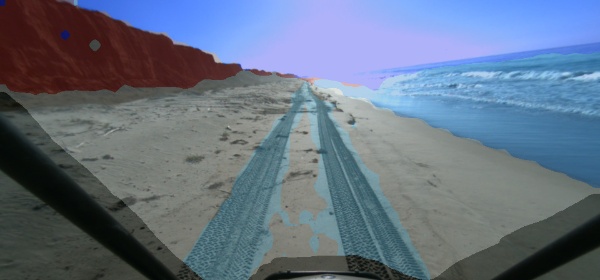}
    \includegraphics[width=0.24\textwidth]{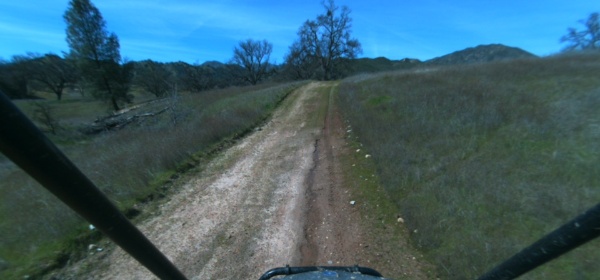} 
    \includegraphics[width=0.24\textwidth]{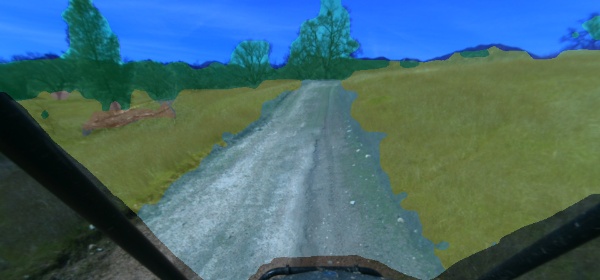}
    %
    %
    \includegraphics[width=0.24\textwidth]{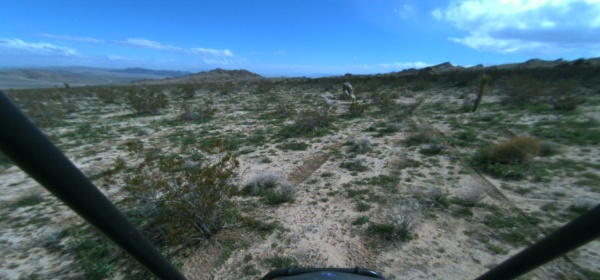}
    \includegraphics[width=0.24\textwidth]{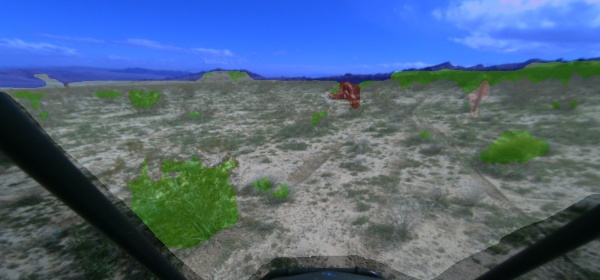}
    %
    \includegraphics[width=0.24\textwidth]{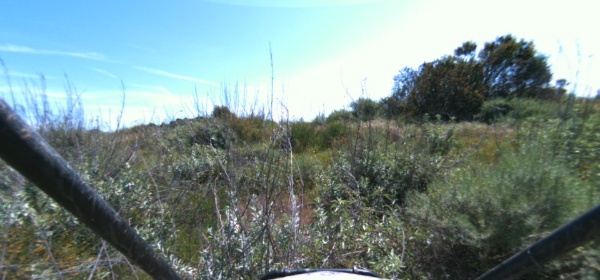} 
    \includegraphics[width=0.24\textwidth]{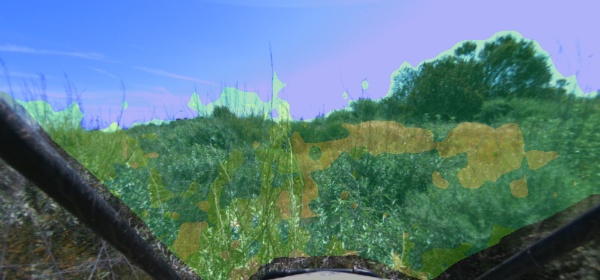}
    %
    \includegraphics[width=0.24\textwidth]{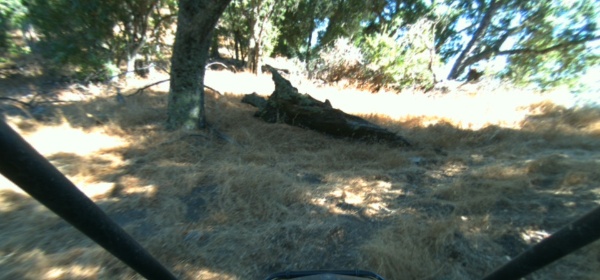}
    \includegraphics[width=0.24\textwidth]{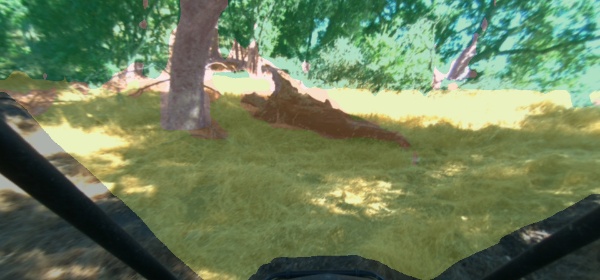}
    %

    \includegraphics[width=1.0\textwidth]{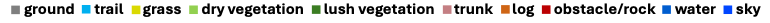}
    
    \caption{Figure of multi-biome sample images (left \& middle right columns) and corresponding semantic prediction overlays (middle left \& right columns). There is a river crossing, beach driving, trails with a log hazard in the grass, desert with rocks and a cactus, dense vegetation, and under-canopy driving scenarios.}
    \label{fig:qual_racer_2d}
\end{figure*}

%
%
\bibliographystyle{named}
\bibliography{project}

\section{Appendix}
The body of this work used the Yamaha and the Rellis datasets since they are open-source.
This entailed performing a class remapping, which simplified the class structure.
Therefore we provide qualitative samples in Figure \ref{fig:qual_racer_2d} from our test sites with the full 10 class predictions using a model trained on our multi-biome dataset in Table \ref{tab:dataset}.
The training hyperparameters for the semantic segmentation models are as follows: batch size of 16, maximum epochs of 110, learning rate of 0.001, and weight decay of 0. The data augmentation parameters include a random cropping and flipping probability of 0.5, a resizing scale of (1440, 900), and a ratio of (0.5, 0.2).

%
\end{document}